\documentclass[10pt, a4paper]{article}

\usepackage[final]{lrec2026} 

\usepackage{latexsym}
\usepackage{booktabs}

\usepackage{fontspec} 
\usepackage{fontspec}
\usepackage{xcolor}
\usepackage{colortbl}
\usepackage{array}
\usepackage{url}
\usepackage{hyperref}
\usepackage{graphicx}
\usepackage{tabularx} 
\usepackage{graphicx}
\usepackage{xcolor}         
\usepackage{comment}
\usepackage{multirow}
\usepackage{xspace}
\usepackage{subcaption}
\usepackage{pifont}
\usepackage{ragged2e}
\usepackage{enumitem}
 \usepackage{amsmath}
\usepackage{arydshln} 
\usepackage{cleveref} 
\usepackage{xcolor}

\definecolor{boxblue}{HTML}{C6E2FF}
\definecolor{boxorange}{HTML}{FFD8B5}
\definecolor{boxgreen}{HTML}{D4EDDA}

\newcommand*{\yoruba}{Yor\`ub\'a\xspace}
\newcommand*{\yoner}{\textsc{YoNER}\xspace}

\title{YoNER: Development of a Yoruba Multidomain Named Entity Dataset}
\title{YoNER: A New Yor\`ub\'a Multi-domain Named Entity Recognition Dataset}

\name{\normalsize Peace Busola Falola$^{1}$, Jesujoba O. Alabi$^{2}$, Solomon O. Akinola$^{1}$, Folashade T. Ogunajo$^{3}$, \\
\bf \normalsize Emmanuel Oluwadunsin Alabi$^{1}$, David Ifeoluwa Adelani$^{5}$}

\address{\footnotesize $^1$University of Ibadan, Nigeria, $^2$Saarland University, Germany, $^3$Atiba University, Nigeria, \\
\footnotesize
 $^5$Mila - Quebec AI Institute, McGill University, and Canada CIFAR AI Chair \\
         peacefalola@gmail.com, david.adelani@mcgill.ca \\
         }

\abstract{
Named Entity Recognition (NER) is a foundational NLP task, yet research in \yoruba has been constrained by limited and domain-specific resources. Existing resources, such as MasakhaNER (a manually annotated news-domain corpus) and WikiAnn (automatically created from Wikipedia), are valuable but restricted in domain coverage. To address this gap, we present YoNER, a new multidomain \yoruba NER dataset that extends entity coverage beyond news and Wikipedia. The dataset comprises about 5,000 sentences and 100,000 tokens collected from five domains including Bible, Blogs,
Movies, Radio broadcast and Wikipedia, and annotated with three entity types: Person (PER), Organization (ORG) and Location (LOC), following CoNLL-style guidelines. Annotation was conducted manually by three native \yoruba speakers, with an inter-annotator agreement of over 0.70, ensuring high quality and consistency. We benchmark several transformer encoder models using cross-domain experiments with MasakhaNER 2.0, and we also assess the effect of few-shot in-domain data using \yoner and cross-lingual setups with English datasets. Our results show that African-centric models outperform general multilingual models for \yoruba, but cross-domain performance drops substantially, particularly for blogs and movie domains. Furthermore, we observed that closely related formal domains, such as news and Wikipedia, transfer more effectively. In addition, we introduce a new \yoruba-specific language model (OyoBERT) that outperforms multilingual models in in-domain evaluation. We publicly release the YoNER dataset and pretrained OyoBERT models to support future research on \yoruba natural language processing.
 \\ \newline  }

\begin{document}

\maketitleabstract

\section{Introduction}

Named Entity Recognition (NER) is a foundational NLP task that identifies and classifies named entities (e.g. personal name, organizations, locations) in text with several applications in information extraction, question answering, and speech recognition~\citep{tjong-kim-sang-de-meulder-2003-introduction,yamada-etal-2020-luke,caubriere-etal-2020-named}. To enable broad applicability, NER models are developed for diverse domains~\citep{weischedel2011ontonotes}, and techniques to facilitate faster adaptation across multiple domains remain an active area of research~\citep{yang2020simple,das2022container,ashok2023promptner,xue2024robust,huang2025adversity}.  
While there are several \textit{multi-domain} datasets available for high-resource languages such as English, they are often lacking for low-resource languages, where existing datasets are typically limited to Wikipedia or news domains~\citep{palen2024openner}.

For low-resource languages such as Yorùbá, spoken by over 50 million people in Nigeria and neighboring African countries, NER is crucial for information extraction and also play an important function of recognizing African names and cultural concepts which are often unrecognized by current AI systems~\citep{olatunji2023afrinames}. Recently, there have been some efforts to create human-annotated NER datasets for \yoruba but they are mostly in the ``\textsc{news}'' domain~\citep{alabi2020massive,adelani2021masakhane,adelani-etal-2022-masakhaner}. Outside \textsc{news}, the other available data is WikiANN based on Wikipedia, however, the annotations are automatically created and only 100 test samples and few entities.


In this paper, we develop the first multi-domain \yoruba NER dataset known as \textbf{\textsc{\yoner}}. \yoner covers five distinct domains including: \textit{Bible}, \textit{Blogs}, \textit{Movies}, \textit{Radio} broadcast, and \textit{Wikipedia}. Each domain has between 800 and 1400 sentences collected from various sources by a \yoruba native speaker. We then recruited three native speakers for the annotation and quality control. Leveraging \yoner, we attempt to answer the following research questions: (1) How well does \textsc{News} NER data transfer to diverse domains? (2) How well can small in-domain data e.g. $200$ sentences improve domain transfer? (3) Does monolingual BERT models enhance multi-domain transfer compared to multilingual BERT models in low-resource setting? To answer the third question, we pre-trained a new \yoruba BERT model (\textsc{Oyo-BERT}) and compare the performance to African-centric BERT models such as AfroXLMR~\cite{alabi2022adapting}. Finally, we extend this evaluation to English language to see if the same result holds true, by comparing English RoBERTa-Large with XLM-Roberta-Large.

Our evaluation shows adapting NER model to other domains is quite challenging especially for the \textsc{Blog} and \textsc{Movie} domain, often with fewer entities. The best adaptation was from \textsc{News} to \textsc{Wikipedia}. Moreover, adapting from \yoruba \textsc{News} domain achieved better performance than English CoNLL NER data with almost twice of its training data, this shows that both source language and closeness of source domains to target are equally important.  We obtained a better adaptation by combining the small in-domain data from all sources and the \textsc{News} domain to achieve the most impressive results. While the newly developed OyoBERT seem to achieve better performance than AfroXLMR, it does not seem to achieve  better cross-domain transfer results.\footnote{We release YoNER publicly on \href{https://huggingface.co/datasets/Peasomama/MultidomainYorubaNER}{Hugging Face}.}\footnote{We release OyoBERT models publicly: \href{https://huggingface.co/Davlan/oyo-bert-base}{YoBERT-base}, \href{https://huggingface.co/Davlan/oyo-mt-bert-base}{OyoBERT-base}, \href{https://huggingface.co/Davlan/oyo-mt-bert-large}{OyoBERT-large}.} 






\begin{table*}[!ht]
\centering
\begin{tabular}{l|p{13cm}}
\toprule
\textbf{Domain} & \textbf{\yoruba Sentence} \\
\midrule
Bible & Àwọn ọmọkùnrin 
\colorbox{boxorange}{Simeoni}, 
\colorbox{boxorange}{Jemueli}, 
\colorbox{boxorange}{Jamini}, 
\colorbox{boxorange}{Ohadi}, 
\colorbox{boxorange}{Jakini}, 
\colorbox{boxorange}{Sohari}, 
\colorbox{boxorange}{Saulu}, 
tí ìyá rẹ jẹ́ ọmọbìnrin ará 
\colorbox{boxgreen}{Kenaani}. \\
\midrule
Blog & Ipò obinrin kò pin si idi àdìrò àti inú ilé yòkù gẹ́gẹ́ bi Olórí Òṣèlú 
\colorbox{boxgreen}{Nigeria}, 
\colorbox{boxorange}{Muhammadu Buhari} ti sọ. \\
\midrule
Movie & Wòó \colorbox{boxorange}{Àṣàké}, jẹ́ ká nasẹ̀ lọ, ká lọ gba'tẹ́gùn díẹ̀. \\
\midrule
Radio & Ìràwọ̀ agbábọọ̀lù tìnú ẹgbẹ́ ìkọ 
\colorbox{boxblue}{Liverpool} 
\colorbox{boxorange}{Salah} nló tí ńlérí wí pé ẹgbẹ́ agbábọọ̀lù náà yóò gba ìfẹ ẹ̀yẹ ní sáà tó ńbọ̀. \\
\midrule
Wikipedia & Òun ló gba àmì ẹ̀yẹ
\colorbox {boxorange}{PEN Pinter} Prize ní ọdún 2018 Iléẹ̀kọ́ sẹ́kọ́ndìrì 
\colorbox{boxblue}{Yunifásítì ti Naijiria}, 
\colorbox {boxgreen}{Nsukka} ni 
\colorbox {boxorange}{Adichie} ti parí ẹ̀kọ́ sẹ́kọ́ndìrì rẹ̀ níbi tí ó ti ọ̀pọ̀lọpọ̀ àmì ẹ̀yẹ ajemọ́ akadá. \\
\bottomrule
\end{tabular}
\caption{Examples of named entities across sentences from different \yoruba domains in YoNER. (\colorbox{boxorange}{PER}, \colorbox{boxblue}{ORG}, \colorbox{boxgreen}{LOC}).}
\label{tab:yoruba_entity_examples}
\end{table*}

\section{Related Work}
\label{sec:append-how-prod}

NER for low-resource languages has recently gained growing attention as the development of multilingual models such as mBERT~\citeplanguageresource{devlin-etal-2019-bert} and XLM-R~\citeplanguageresource{conneau-etal-2020-unsupervised} has enabled effective transfer learning across languages. Traditional NER systems relied heavily on large annotated datasets, which are scarce for most low-resource languages. Multilingual pre-trained language models have helped bridge this gap by transferring knowledge from high-resource languages to creating competitive models in zero and few-shot setting~\citep{zhao-etal-2021-closer,adelani2021masakhane,schmidt-etal-2022-dont}. This progress has opened new opportunities for languages that were previously under-served in NLP research.

In the African context with several low-resource languages, significant progress has been achieved through community-driven initiatives aimed at building foundational datasets and benchmarks. The first effort was by \citet{alabi2020massive} that developed \yoruba NER data based on Global Voices news articles, with 26K tokens. The MasakhaNER project (MasakhaNER 1.0 and MasakhaNER 2.0) created larger resources~\citeplanguageresource{adelani-etal-2021-masakhaner,adelani-etal-2022-masakhaner} with 83K and 244K tokens (between 3K and 10K sentences)  respectively. 
The MasakhaNER datasets represents high-quality, manually annotated NER datasets for over 21 African languages from the news domain, including \yoruba. Other NER datasets available are WikiAnn~\citeplanguageresource{pan-etal-2017-cross} and NaijaNER~\citeplanguageresource{oyewusi2021naijaner}. The former is automatically annotated from Wikipedia by projecting annotations from English, however the size is small (100 sentences test set) while the latter is also based on news domain but they expanded the entity types from four (PER, ORG, LOC and DATE) to 18 entity types covered by OntoNotes~\citep{weischedel2011ontonotes}. Existing datasets do not cover diverse domains, and we address this gap in this paper. 

Cross-domain NER is an active research area, as models trained on one genre typically perform poorly on others due to shifts in vocabulary and entity distributions~\citep{jia-etal-2019-cross,liu2021crossner}. While there have been efforts to develop new methods and datasets to address this issue~\citep{jia-etal-2019-cross}, most of these have focused on high-resource languages such as English, leaving low-resource languages relatively underexplored and underrepresented in cross-domain NER research.


Our work contributes to this line by providing multi-domain human annotated data for \yoruba called \textbf{\yoner}. Given \textbf{\yoner}, we conducted several experiments to study the cross-domain transfer across the domains using several transformer based language models. 

\section{Corpus Curation for \yoner}

\subsection{Data Collection}

We compiled \yoruba texts from five domains, aiming for varied style and content. We focused on popular domains covered in previous works such as OntoNotes and WikiAnn: \textit{Bible}, \textit{Blog}, \textit{Radio}, \textit{Wikipedia}, \textit{Movie}. For the two domains, movies and radio, with which we were less familiar, we collaborated with a Yoruba newscaster to collect sentences. More details about the data collection process are provided below. 

\textbf{Bible}: We collected verses from the \yoruba Bible (Bible.com corpora). Based on the authors familiarity with the Bible, we chose books and chapters that contained more entities such as \texttt{Genesis 10}, \texttt{Genesis 14}, \texttt{Exodus 6}, \texttt{Numbers 1}, \texttt{Numbers 3}, \texttt{Numbers 26}, \texttt{Joshua 12}, \texttt{Joshua 18}, \texttt{I Samuel 8:1}, \texttt{I Samuel 22}, \texttt{II Samuel 5}, \texttt{I Kings 9}, \texttt{Nehemiah 3}, \texttt{Ezra 2:1}, \texttt{Psalms 87}, \texttt{Jeremiah 25}, \texttt{Joel 3}, \texttt{Matthew 2}, \texttt{Matthew 23}, \texttt{Luke 2}, \texttt{Acts 6}, \texttt{Acts 15} and \texttt{Acts 27}. 

\textbf{Blogs}: We curated posts and comments from \yoruba blogs and forums (e.g. Nairaland, social media pages, and Yoruba news blogs). This domain includes informal language, slang, and possible code-mixing. 

\textbf{Movies}: We collected transcripts of dialogues from \yoruba movies (collected from YouTube subtitled content). Movie dialogue often contains conversational \yoruba, interjections, and may omit diacritics. Sentences were gathered from four movies for the movies domain. The movies are \texttt{Oleku}, \texttt{Apoti Eri}, \texttt{Jagunjagun} and \texttt{Owo Eje}, all obtained from YouTube and Netflix. All movies except Apoti Eri was code-mixed. A \yoruba newscaster manually transcribed all the movies. 

\textbf{Radio}: We collected transcripts and recordings from radio stations broadcasting in \yoruba. 
(e.g. Tiwa-n-Tiwa, Fresh FM). These include news and talk shows, blending formal and colloquial speech.

\textbf{Wikipedia (Wiki)}: We make use of \yoruba Wikipedia articles obtained from HuggingFace.\footnote{\url{https://huggingface.co/datasets/wikimedia/wikipedia}} This domain is encyclopedic and covers a broad range of topics but uses a formal written style.

\autoref{tab:yoruba_entity_examples} presents an example sentence for each domain covered in \yoner, along with the corresponding annotations. The examples include annotated personal names and locations from different domains, such as a list of biblical names in the Bible domain and the mention of a Nigerian political figure in a blog.


\paragraph{Data split} 
After cleaning and sentence segmentation of the collected data, the sentences were divided into splits with final counts of 200 for training, 100 for development, and the remaining sentences for testing (ranging from ~500 to ~1,128 test sentences per domain). In total, the corpus contains 5,148 sentences (100,795 tokens) across domains. \autoref{tab:yoruba_dataset_stats} shows the statistics of the dataset.

\begin{table}[t]
\centering
\small 
\scalebox{0.78}{
\begin{tabular}{l|p{2.6cm}ll}
\toprule
\textbf{Domain} & \textbf{Source} & \textbf{Train/dev/test} & \textbf{Tokens} \\
\midrule
\multicolumn{4}{l}{\texttt{YoNER}}\\

Blogs & Nairaland, Yoruba blog, Akonilekede Yoruba, Facebook & 200/100/553 & 20,123 \\
\midrule
Bible & Bible.com & 200/100/713 & 23,896 \\
\midrule
Movies & YouTube & 200/100/1128 & 14,312 \\
\midrule
Radio & Tiwa-n-Tiwa and Fresh FM Radio Stations & 200/100/752 & 21,404 \\
\midrule
Wikipedia & Wikipedia & 200/100/502 & 21,090 \\
\midrule
\textbf{Total} &  & \textbf{5,148} & \textbf{100,795} \\
\midrule
\multicolumn{4}{l}{\texttt{Existing dataset}}\\
News & MasakhaNER 2.0 & 6877/983/1964  & 244,144 \\
\bottomrule
\end{tabular}
}
\caption{\yoner Dataset statistics by domain.}
\label{tab:yoruba_dataset_stats}
\end{table}

 \subsection{Annotation Process}

We recruited three native \yoruba speakers (who are also authors on this paper) to annotate the collected sentences of various domains. Annotation was done via the Human Signal annotation platform (formerly Label Studio). Annotators were trained on the NER annotation guidelines used by MasakhaNER~\cite{adelani2021masakhane}  to label each token for three entity types: PER (person), LOC (location), and ORG (organization). We used the IOB2 scheme for spans: each token is tagged with B-, I-, or O (outside any entity).

Each sentence was annotated independently by all three annotators. 
\paragraph{Inter-Annotator Agreement} After labeling, we computed inter-annotator agreement using Fleiss’ Kappa at two levels: token level (ignoring B/I differences, just whether each token is labeled as an entity or not) and entity level (treating each contiguous span as an entity annotation). The results in \Cref{tab:fleiss_kappa_results} show substantial agreement at the token level (k=0.71–0.88) but much lower at the entity level (k=0.30–0.61). This gap indicates that annotators often agreed on which words are entities but had more disagreement on exact span boundaries (a common issue in NER). Disagreements were resolved by majority vote (or discussion when all three annotators differed) to produce the final gold-standard labels.

The final annotated dataset is therefore high-quality Yoruba NER data across five domains. \autoref{fig:entity_types} shows the entity counts across each domain. Across all domains, PER entities are the most frequent, while ORG entities are rare, appearing mainly in data from Radio and Wikipedia.



\begin{table}[t]
\centering
\small
\setlength{\tabcolsep}{7pt} 
\renewcommand{\arraystretch}{1.25} 
\begin{tabular}{|l|r|r|}
\hline
\textbf{Domain} & \textbf{Fleiss’ $\kappa$ (token)} & \textbf{Fleiss’ $\kappa$ (entity)} \\
\hline
Blogs & 0.7064 & 0.3957 \\
\hline
Bible & 0.8775 & 0.5938 \\
\hline
Movies & 0.7911 & 0.3038 \\
\hline
Radio & 0.8313 & 0.6072 \\
\hline
Wikipedia & 0.8190 & 0.5758 \\
\hline
\end{tabular}
\caption{Inter-annotator agreement (Fleiss’ $\kappa$) across Yoruba NER domains.}
\label{tab:fleiss_kappa_results}
\end{table}

\begin{table*}[ht]
\centering
\resizebox{\linewidth}{!}{
\begin{tabular}{lcrrrrrrr}
\toprule
 & & \textbf{Topic} & \textbf{News Topic} & \textbf{Twitter} & \textbf{Movie} &  &  &   \\
\textbf{Model} & \textbf{Size} & \textbf{Class.} & \textbf{Class.} & \textbf{Sentiment} & \textbf{Sentiment} &  \textbf{NER} & \textbf{POS} & \textbf{Avg.}  \\
\midrule
\multicolumn{8}{l}{\texttt{Baselines}} \\
AfriBERTa~\cite{ogueji-etal-2021-small} & 126M & 70.6 & 90.3  & 72.9 & 90.9 & 87.7 & 94.5 & 84.5\\
AfroXLMR-large~\citep{alabi2022adapting} & 550M & 74.8 & 94.0 & 74.1 & 84.3 &  \textbf{89.3} & \textbf{95.0} & 85.3 \\
AfroXLMR-large-76L~\citep{adelani-etal-2024-sib} & 550M & 79.9 & \textbf{94.7} & \textbf{75.1} & 88.0 & 88.4 & 95.2 & 86.8 \\
\midrule
YoBERT-base (ours) & 110M & 69.2 & 92.1 & 71.1 & 90.0 & 85.0 & 94.0 & 83.6 \\
OyoBERT-base (ours) & 110M & 80.5 & 94.0 & 73.0 & 90.6 & 86.7 & 94.4 & 86.5 \\
OyoBERT-large (ours) & 337M &  \textbf{82.5} & 93.8 & 74.9 & \textbf{92.6} & 87.4 & 94.6 & \textbf{87.6} \\
\bottomrule
\end{tabular}
}
\caption{Evaluation of OyoBERT models and other multilingual baselines on four sequence-level and two token-level classification tasks for \yoruba.}
\label{tab:oyo_bert}
\end{table*}

\section{Experimental Setup}
Given the created \yoner dataset, in this section we outline our experimental setup designed to answer the research questions posed earlier.

\subsection{Cross-Domain Transfer of News NER}
To address the first research question, “\textit{How well does News NER data transfer to diverse domains?},” we trained NER models on data from the News domain and evaluated them on other domains. Specifically, we fine-tuned several multilingual transformer models on the \yoruba portion of the MasakhaNER 2.0~\citeplanguageresource{adelani-etal-2022-masakhaner} dataset, which primarily consists of news text. We selected MasakhaNER 2.0 because of its moderately large size (over 6,000 training sentences) and its rich collection of African geographical and cultural entities, making it a suitable source domain for evaluating cross-domain transfer in \yoruba. The resulting models were then evaluated on the test splits of each domain in the \yoner dataset.
\paragraph{Models:} The selected multilingual PLMs include AfriBERTa~\citeplanguageresource{ogueji-etal-2021-small}, AfriBERTa-V2~\citeplanguageresource{oladipo2024scaling}, mBERT~\citeplanguageresource{devlin-etal-2019-bert}, XLM-R (base and large;~\citeplanguageresource{conneau-etal-2020-unsupervised}), and AfroXLMR (base, large, and large-76L;~\citeplanguageresource{alabi2022adapting,adelani-etal-2024-sib}). 

\subsection{Impact of Small In-Domain Data on NER Transfer}

To address the second research question, “\textit{How well can small in-domain data (e.g., 200 sentences) improve domain transfer?},” we fine-tuned the best-performing base model from the previous experiment on the training split of each domain in \yoner. The resulting models were then evaluated both in-domain and out-of-domain to assess how a small amount of domain-specific data affects NER performance across diverse text domains. To provide a cross-lingual baseline for comparison, we also included English NER datasets covering domains similar to those in \yoner. Specifically, we used CoNLL and WikiAnn~\citeplanguageresource{pan-etal-2017-cross} for the News and Wikipedia domains, respectively, while from the OntoNotes~\citep{weischedel2011ontonotes} dataset, we selected broadcast conversations and web blogs to represent the Radio and Blogs domains. In this case, all available English data for each domain were used for training, and the Bible domain was included only in the evaluation phase, as no corresponding English training data were available. Furthermore, all the English datasets were processed to have only PER, LOC and ORG as entity types, all other entity types were changed to O except for OntoNotes where we changed GPE to LOC.

\begin{figure}[t]
\begin{center}
\includegraphics[width=\columnwidth]{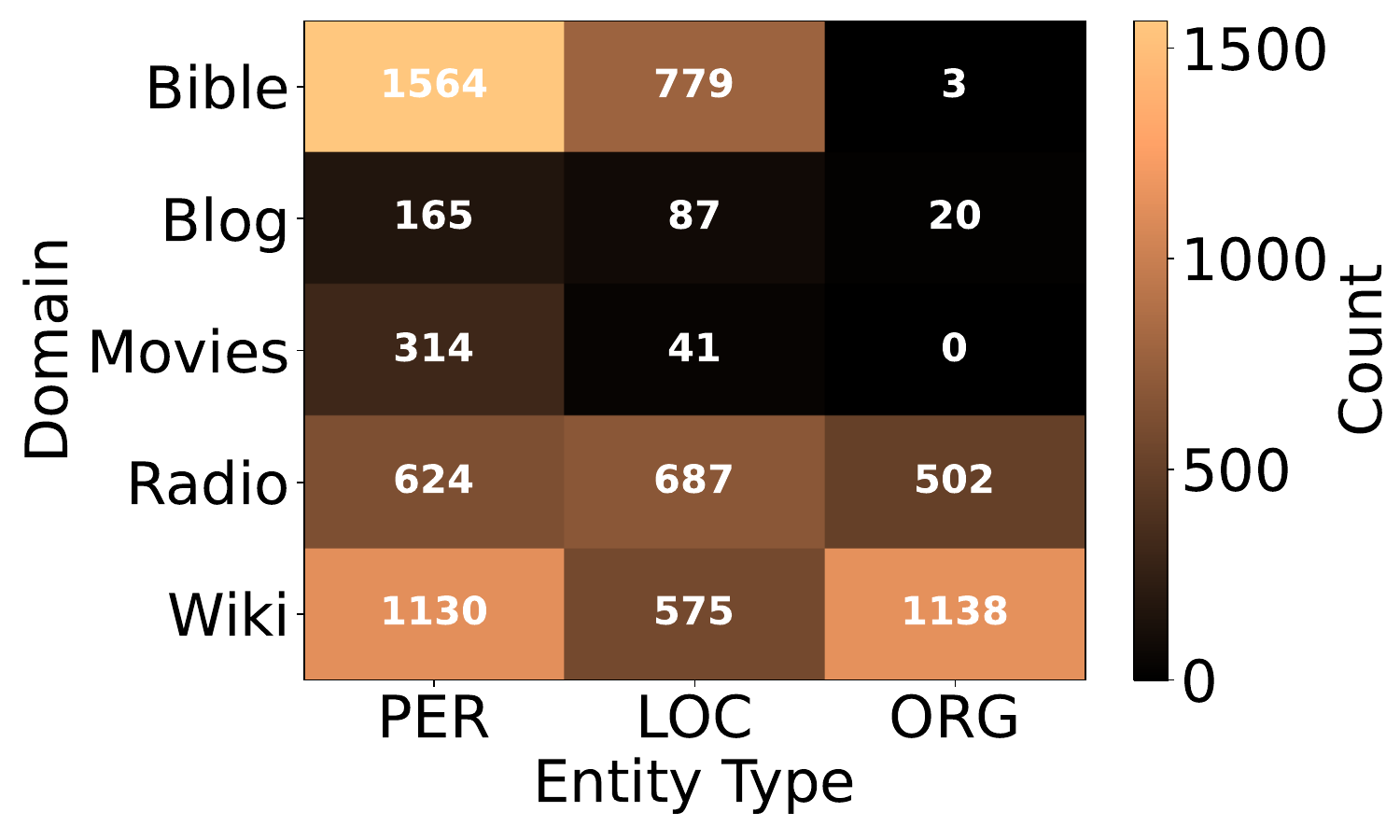}
\caption{Frequency of Entity Types By Domain in \yoner dataset.}
\label{fig:entity_types}
\end{center}
\end{figure}

\subsection{Monolingual vs. Multilingual Models for Multi-Domain Transfer}
Previous research has shown that multilingual transformer models often underperform compared to their monolingual counterparts across different languages~\citep{ronnqvist-etal-2019-multilingual,pyysalo-etal-2021-wikibert}. We investigate this observation in our third research question. To address the question, “\textit{Do monolingual BERT models enhance multi-domain transfer compared to multilingual BERT models?},” we compare Named Entity Recognition (NER) models trained in English using RoBERTa (a monolingual model) with XLM-R (a multilingual model). For \yoruba, since no well-established monolingual models are available but multilingual models exist, we develop a BERT model for \yoruba, as described in \Cref{sec:oyo_bert}. We fine-tuned the BERT model on each individual domain and evaluated it on the corresponding test set, comparing the results to fine-tuning on a combination of all five domains simultaneously. For the \yoner experiments, we compared BERT with AfroXLMR-large-76L under this setup. We replicated the same procedure for English as well. To replicate a similar low-resource setting as in \yoner, where we had only 200 training examples per domain, our initial experiments on the English datasets showed that approximately 1,000 examples and relatively long sentences were needed to achieve reasonable in-domain F1 scores. Therefore, for the English experiment, we randomly selected 1,000 sentences from CoNLL, WikiAnn, and OntoNotes (as described in the previous section) containing more than five tokens, and compared RoBERTa-large with XLMR-large.

\subsection{Oyo-BERT pre-training}
\label{sec:oyo_bert}

To address the third question, we pre-trained a BERT model for \yoruba. 
We trained a BERT model with token dropping objective~\citep{hou-etal-2022-token}---where unimportant tokens are dropped in the intermediate layer but later picked up by the last layer so that the model still produces full-length sequences. We make use of an  open-source implementation.\footnote{\url{https://github.com/stefan-it/model-garden-lms}} The BERT model was trained on a TPU v3-8 Google cloud compute in less than 24 hours. 

The pre-training data is based on the \yoruba subset of mC4---pre-training corpus for mT5~\citep{xue-etal-2021-mt5} (153MB), MT560~\citep{gowda-etal-2021-many} (59MB), \yoruba portion of the Wikipedia (11MB), and a number of curated news sources such as BBC \yoruba (15MB), Alaroye (10MB), Awinkoko (7MB), and Asejere (2MB). The total size of the monolingual data was around 347MB. 

To further increase training data, we leverage synthetic data obtained by machine translating additional contents for English, similar to how AfroXLMR-76L was created~\cite{adelani-etal-2024-sib}--where languages less than 10MB leverage MT generated data from NLLB-200 (600M parameters)~\cite{costa2022no}.\footnote{Important to note that \yoruba did not use synthetic data in AfroXLMR since it has more than 300MB}. Here, we translated over 1GB of English texts from WURA~\citep{oladipo-etal-2023-better}---An high-quality African corpus containing English to \yoruba with NLLB (600M)~\citep{nllb2024scaling}. This increased the entire pre-training corpus to over 1.93GB. 

The resulting models are trained for two model sizes: The \textbf{OyoBERT-base} has the same configuration as BERT-base while the \textbf{OyoBERT-Large} has same configuration as BERT-large. The BERT trained without MT data is referred to as \textbf{YoBERT}.

\begin{table*}[ht]
\centering
\scalebox{0.90}{
\begin{tabular}{lcccccccc}
\toprule
\textbf{Model} & \textbf{Bible} & \textbf{Blog} & \textbf{Movies} & \textbf{Radio} &  \textbf{Wiki} & \textbf{News}  & \textbf{Avg.}  \\
\midrule
\multicolumn{8}{l}{MasakhaNER 2.0} \\
OyoBERT-large & 69.80$_{1.10}$ & 54.30$_{1.23}$ & 42.17$_{1.53}$ & 72.50$_{3.03}$ & 74.14$_{1.31}$ & 87.78$_{0.49}$ & 66.78 \\
AfriBERTa & 58.85$_{1.72}$ & 52.77$_{1.18}$ & 44.47$_{1.76}$ & 75.26$_{0.72}$ & 68.96$_{0.95}$ & 87.42$_{0.58}$ & 64.62 \\
AfriBERTa-V2 & 71.01$_{1.22}$ & 50.38$_{2.10}$ & 47.12$_{1.84}$ & \textbf{76.47}$_{2.11}$ & 75.01$_{0.45}$ & 87.22$_{0.29}$ & 67.87 \\
mBERT & 61.94$_{1.54}$ & 43.38$_{3.71}$ & 25.72$_{4.34}$ & 76.10$_{1.22}$ & 70.10$_{0.63}$ & 84.30$_{0.35}$ & 60.26 \\
XLM-R-base & 63.20$_{1.21}$ & 40.22$_{1.54}$ & 42.40$_{3.47}$ & 68.54$_{1.52}$ & 66.28$_{1.18}$ & 84.19$_{0.51}$ & 60.81 \\
XLM-R-large & 68.80$_{2.24}$ & 42.24$_{2.26}$ & 42.19$_{5.03}$ & 68.38$_{2.19}$ & 67.62$_{1.86}$ & 84.12$_{0.88}$ & 62.23 \\
AfroXLMR-base  & 73.90$_{1.58}$ & 50.71$_{1.56}$ & 39.63$_{2.63}$ & 72.63$_{1.64}$ & 73.88$_{1.03}$ & 86.59$_{0.32}$ & 66.22 \\
AfroXLMR-large & 74.48$_{0.72}$ & 58.50$_{1.17}$ & 49.19$_{5.83}$ & 71.36$_{1.23}$ & 77.35$_{0.80}$ & \textbf{88.77}$_{0.42}$ & 69.94 \\
AfroXLMR-large-76L & \textbf{77.22}$_{2.39}$ & \textbf{59.07}$_{3.33}$ & \textbf{50.69}$_{2.61}$ & 70.49$_{2.06}$ & \textbf{77.88}$_{1.91}$ & 88.04$_{0.73}$ & \textbf{70.57} \\

\bottomrule
\end{tabular}
}
\caption{F1 score (\%) for cross-domain transfer to the YoNER dataset from the News domain. Scores are averaged over five runs. All models are fine-tuned on the training split of MasakhaNER 2.0.}
\label{tab:newscrossdomain}
\end{table*}

\begin{figure*}[ht]
    \centering
    \begin{subfigure}[t]{0.48\textwidth}
        \centering
        \includegraphics[width=\textwidth]{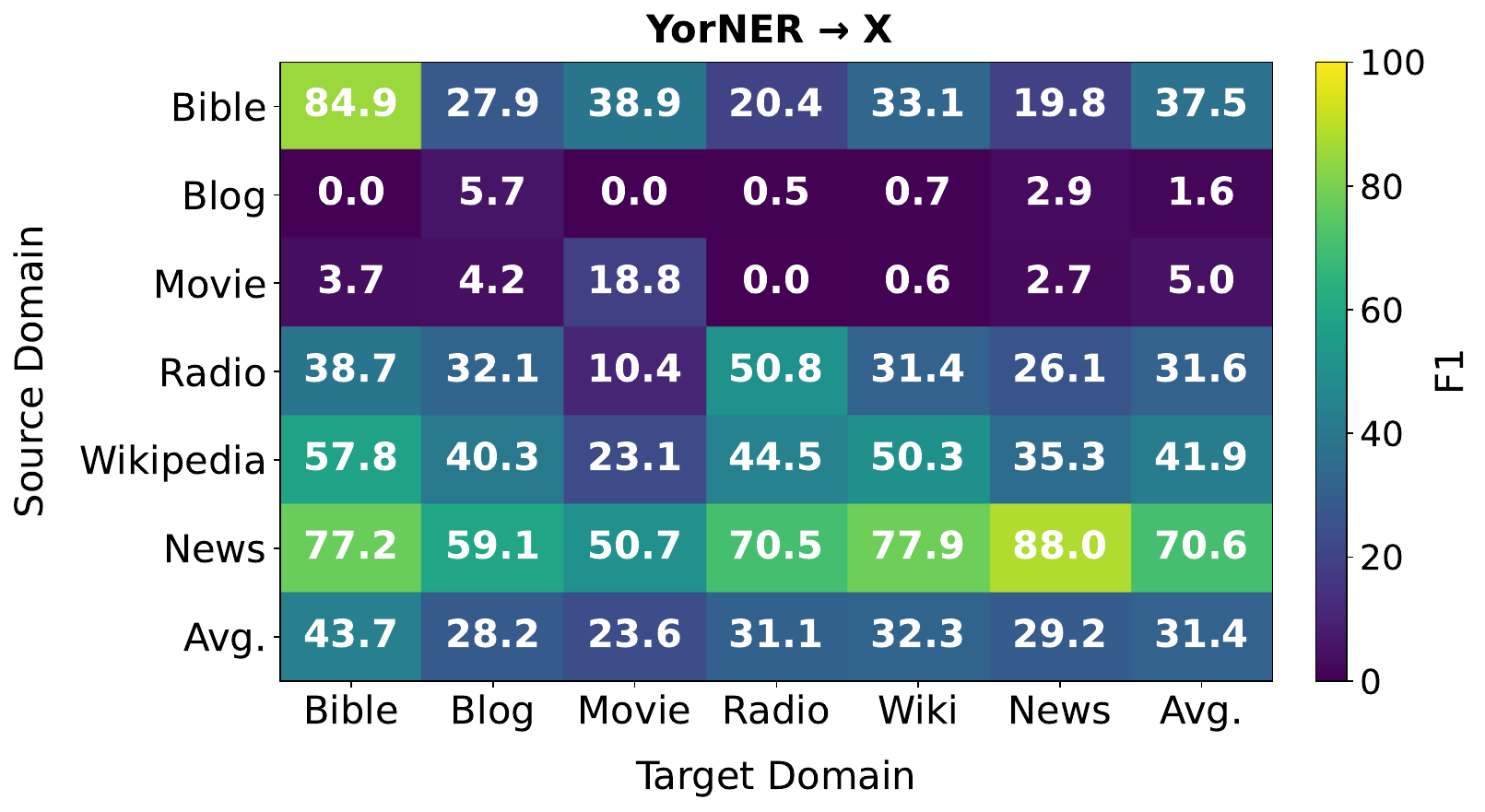}
        \caption{\yoner → X}
        \label{fig:yornertransf}
    \end{subfigure}
    \hfill
    \begin{subfigure}[t]{0.48\textwidth}
        \centering
        \includegraphics[width=\textwidth]{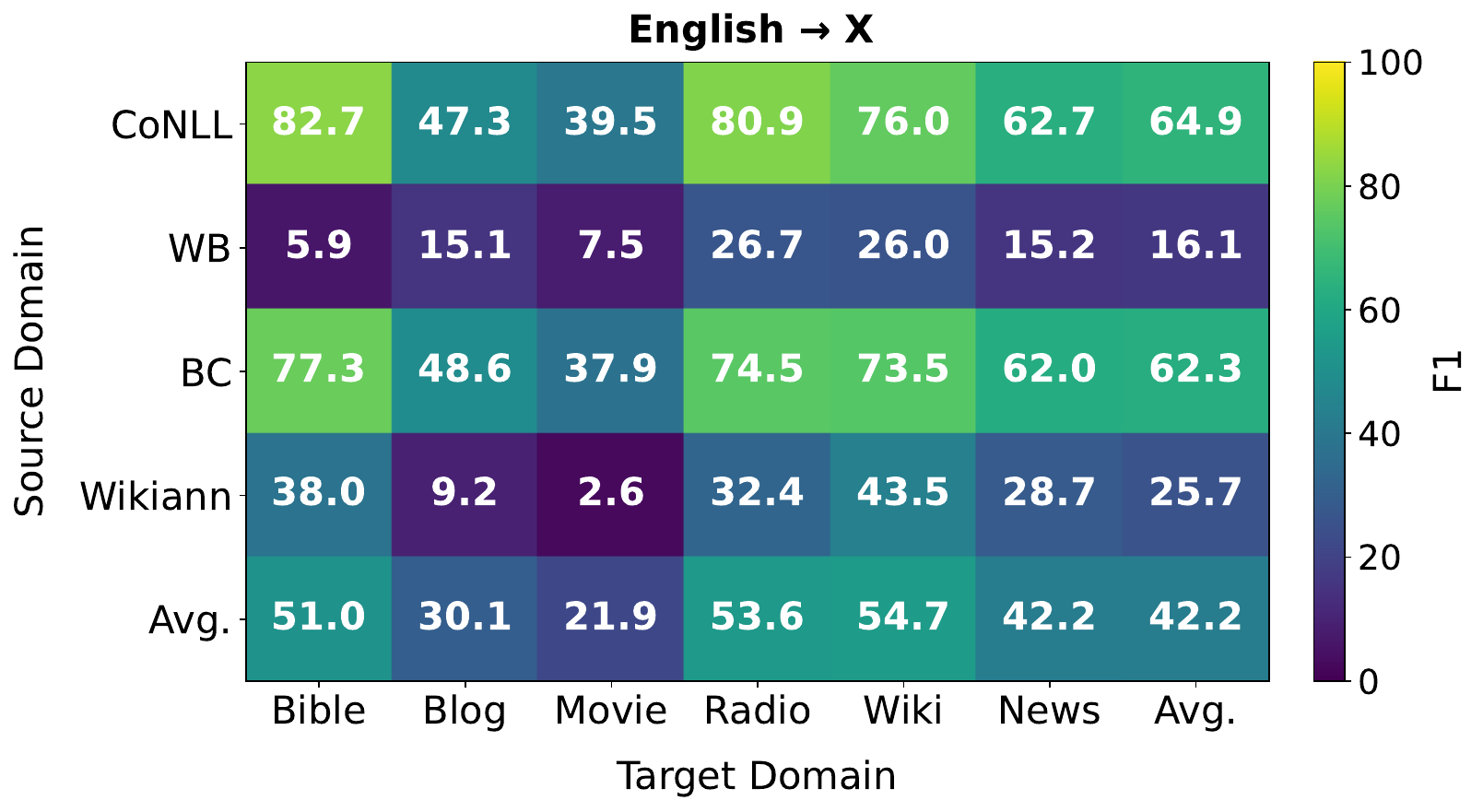}
        \caption{English → X}
        \label{fig:engnertransf}
    \end{subfigure}

    \caption{Cross-domain evaluation of NER model trained on each YoNER domain and on domain-specific English NER datasets. English-trained models are evaluated on YoNER domains to assess cross-lingual transfer. Average over 5 runs.}
    \label{fig:two_heatmaps}
\end{figure*}

\subsection{Model Hyper-parameters}
For all NER fine-tuning experiments, we adapted an open-source codebase.\footnote{\url{https://github.com/masakhane-io/masakhane-ner/tree/main/MasakhaNER2.0}} We used a maximum sequence length of 256, a batch size of 32, a learning rate of 5e-5, and trained for 50 epochs. For evaluation, we report the micro-averaged F1 score.

\section{Result and Discussion}
In this section, we present the results of our experiments, beginning with the evaluation of the newly created PLMs for \yoruba. We then proceed to address the research questions outlined earlier.

\subsection{How well does OyoBERT perform on wide range of tasks?}

Here, we compared the performance of the \textit{newly} trained \yoruba BERT models (YoBERT, OyoBERT-base and OyoBERT-large) to well established multilingual baselines such as AfriBERTa~\cite{ogueji-etal-2021-small}, AfroXLMR~\cite{alabi2022adapting} and AfroXLMR-76L~\cite{adelani2024sib} covering 11, 20 and 76 languages respectively including \yoruba. 

We evaluated on six representative NLP evaluation data including SIB-200~\cite{adelani2024sib} (a topic classification dataset to categorize sentences), MasakhaNEWS~\cite{adelani2023masakhanews} (a news topic classification to categorize news articles), AfriSenti~\cite{muhammad2023afrisenti} (a Twitter sentiment classification dataset), NollySenti~\citep{shode2023nollysenti} (a movie sentiment classification dataset), MasakhaNER~\cite{adelani2021masakhane} (for NER), and MasakhaPOS~\cite{dione2023masakhapos} (for part-of-speech tagging). 

\autoref{tab:oyo_bert} shows the average evaluation of OyoBERT on six tasks. First observation is that it outperform the YoBERT with about $+3$ points, this shows the benefit of more data, specifically synthetic data generated by MT models. Unsurprisingly, the larger OyoBERT-large achieved better results than the OyoBERT-base. When compared to the strong baselines, OyoBERT-base with only 110M parameters performed better than AfriBERTa with 126M parameters. Similarly, OyoBERT-large with only 337M parameters achieved better overall performance to AfroXLMR-{76L} models with 550M parameters. In general, OyoBERT has comparable or better performance on the sequence classification tasks, however, they struggle with token classification task such as NER, where AfroXLMR-large and AfroXLMR-76 achieved $+1.9$ and $+1.0$ better performance. 

\textit{Overall, OyoBERT-large is the first strong monolingual encoder model for \yoruba with comparable performance to other Africa-centric  models but perform slightly worse on NER.}

\subsection{How well does News
NER data transfer to diverse domains?}

\autoref{tab:newscrossdomain} presents the results obtained for all models when fine-tuned on MasakhaNER 2.0 and evaluated on each domain in \yoner, including the MasakhaNER 2.0 test set, denoted as ``\texttt{News}''. Overall, while all models achieved at least 84\% F1 when the \texttt{News} model was evaluated on its corresponding test set, we observed a drop in F1 across all other \yoner domains. In particular, AfroXLMR-large-76L obtained the highest average F1, reaching 88.04\% on News. The next best domains were Bible and Wikipedia, scoring 77.22\% and 77.88\%, respectively, which represents a drop of nearly 11\%. Blog and Movies showed the lowest cross-domain performance from the News model, with F1 scores of 59.07\% and 50.69\%, respectively. This is also consistent with the fact that Blog and Movies are the domains with the fewest annotated entities (as shown in \autoref{fig:entity_types}).

Furthermore, a look at the average performance of all the models shows that, although many are multilingual except OyoBERT, the African-centric models consistently perform better for cross-domain transfer, while mBERT and XLM-R (base and large) are the least performing models. We attribute the low performance of XLM-R to the fact that \yoruba was not part of it's pre-training languages, while for mBERT, the \yoruba portion is based on Wikipedia which is less than 30k articles. 

\textit{Overall, African-centric models outperform general multilingual models for \yoruba, but cross-domain performance drops substantially, especially for Blog and Movies.}

\subsection{ How
well can small in-domain data improve domain transfer?}
\autoref{fig:yornertransf} shows the cross-domain performance of NER models trained on each domain in YoNER, based on AfroXLMR-large-76L, the best model from the previous experiment. The figure also includes results from Table 4 and the evaluation of YoNER models on MasakhaNER 2.0. The diagonal of the figure represents in-domain performance. As observed, Bible has the highest in-domain performance with 84.9\%, followed by Radio and Wikipedia with 50.8\% and 50.3\%, respectively, while Blog and Movies exhibit the lowest in-domain performance.

Regarding cross-domain transfer, aside from News (MasakhaNER 2.0), Wikipedia is the best source domain, followed by Bible and Radio. The best target domain is Bible, followed by Wikipedia and Radio. However, all YoNER domains fail to transfer effectively to the News domain, even though the News model transfers significantly better to these domains. We attribute this to the large training size of MasakhaNER 2.0, which gives the News model a strong both in-domain and cross-domain performance.

\textit{Overall, closely related domains such as News and Wikipedia with formally written texts transfers better to each other.}

\subsection{How well does Cross-lingual transfer compares to cross-domain transfer?}
\autoref{fig:engnertransf} shows the result from the cross-lingual experiment. For cross-lingual transfer from English datasets, CoNLL, WB, and BC did not perform best when transferred to their corresponding \yoruba domains (News, Blog, and Radio), with only WikiAnn transferring best to Wikipedia. Considering cross-lingual cross-domain transfer, on average, CoNLL and BC were the best source domains, while WB got the lowest F1 score. As target domains, Wikipedia and Radio achieved the highest performance, whereas Blog and Movies were the worst-performing targets. Although Bible is not the best target domain on average, it achieved the highest F1 scores specifically from CoNLL and BC as source domains. In several cases, cross-lingual transfer from English datasets outperforms training on just 200 in-domain examples from \yoner, both for in-domain and cross-domain evaluations. There large data for cross-lingual may be more effective than small available in-domain data. 

However, when we compare fine-tuning on large \yoruba news NER data (MasakhaNER 2.0) with 6.8K sentences and English news NER data (CoNLL03) with 14.9K, we found the former setting to be more effective in adapting to other domains, achieving $70.6$ average performance (\autoref{fig:yornertransf}) while English to other domains achieved $64.9$ (+5.7 points improvement). This finding shows that transfer within the same language is easier than from another language, despite the difference in domains. 


\begin{table}[t]
\centering
\scalebox{0.70}{
\begin{tabular}{lcccccc}
\toprule
\textbf{Model} & \textbf{Bible} & \textbf{Blog} & \textbf{Movies} & \textbf{Radio} &  \textbf{Wiki}  & \textbf{Avg.}  \\
\midrule
\multicolumn{7}{l}{\emph{In-domain}} \\
OyoBERT-large & 81.47 & \textbf{7.78} & \textbf{34.06} & \textbf{65.76} & 43.09 & \textbf{46.43} \\
AfroXLMR-large & \textbf{84.85} & 5.67 & 18.78 & 50.82 & \textbf{50.26} & 42.08 \\
\midrule
\multicolumn{7}{l}{\emph{Multi-domain}} \\
OyoBERT-large & 85.00 & 71.53 & \textbf{56.10} & 89.60 & 76.37 & 75.72 \\
AfroXLMR-large & \textbf{89.63} & \textbf{76.09} & 53.72 & \textbf{90.84} & \textbf{79.64} & \textbf{77.98} \\

\bottomrule
\end{tabular}
}
\caption{F1 score (\%) for In-domain evaluation of each of the YoNER domains. Average is over 5 runs.}
\label{tab:yorcross}
\end{table}

\begin{table}[t]
\centering
\scalebox{0.70}{
\begin{tabular}{lccccc}
\toprule
\textbf{Model} & \textbf{CoNLL} & \textbf{WB} & \textbf{BC} & \textbf{Wikiann}   & \textbf{Avg.}  \\
\midrule
\multicolumn{6}{l}{\emph{In-domain (all data)}} \\
RoBERTa-large & 93.14 & \textbf{64.37} & 88.16 & 82.48 & 82.04 \\
XLM-R-large & \textbf{93.70} & 63.12 & \textbf{91.37} & \textbf{82.74} & \textbf{82.73} \\
\midrule
\multicolumn{6}{l}{\emph{In-domain (1000 examples)}} \\
RoBERTa-large & \textbf{74.33} & \textbf{1.79} & \textbf{40.98} & \textbf{54.14} & \textbf{42.81} \\
XLM-R-large & 53.59 & 0.00 & 18.23 & 40.32 & 28.04 \\
\midrule
\multicolumn{6}{l}{\emph{Multi-domain (1000 examples)}} \\
RoBERTa-large & 82.36 & 48.77 & 69.58 & 64.55 & 66.32 \\
XLM-R-large & \textbf{83.52} & \textbf{49.44} & \textbf{74.50} & \textbf{62.91} & \textbf{67.59} \\
\bottomrule
\end{tabular}
}
\caption{F1 score (\%) for In-domain evaluation of  each of the English domains. Average is over 5 runs.}
\label{tab:engcross}
\end{table}

\textit{Overall, cross-domain transfer within the same language is more effective than cross-lingual transfer especially with large training source data.}

\subsection{Does monolingual BERT models enhance multi-domain transfer compared to multilingual BERT models in low-resource setting?}
\label{sec:multidomainexp}

\autoref{tab:yorcross} shows the in-domain and multi-domain performance of the \yoner datasets, comparing OyoBERT-large and AfroXLMR-large-76L. We observed that, in in-domain evaluation, the language-specific model on average outperforms the multilingual model. Looking at the individual in-domain results, OyoBERT outperforms AfroXLMR in culturally specific domains such as Movies and Radio, but is less competitive on Bible and Wiki, which are not culturally specific domains. For the multi-domain evaluation, AfroXLMR appears to benefit from exposure to more data and, on average, outperforms OyoBERT; however, it still underperforms on the Movies domain. This underscores that language-specific models can be competitive in certain contexts, particularly within culturally specific domains.

Replicating this same experiment in English and English datasets, comparing RoBERTa-large and XLM-R-large (as shown in \autoref{tab:engcross}), we observed findings similar to those from \yoner. The results show that, for in-domain evaluation, the language-specific model RoBERTa-large outperforms its multilingual counterpart. However, under the multi-domain setup, XLM-R-large gains an advantage, likely due to its exposure to a broader range of training data.

\textit{Overall, in low-data regime, monolingual BERT is more beneficial for cross-domain transfer, however with more training data, multilingual BERT-based models win.}

\section{Which entity types do models tend to perform poorly on?}
\label{sec:entitylysis}
To answer this question, we examined the entity type F1 scores for the multidomain models described in \Cref{sec:multidomainexp} and presented in \Cref{tab:engcrossent}. Our results show that for both OyoBERT and AfroXLMR-large, the ORG entity type consistently has the lowest F1 score. Furthermore, for the Bible and Movies domains, there are no ORG entities in the dataset, as illustrated in \Cref{fig:entity_types}. We also found that for the Blog and Radio domains, the models perform better at identifying LOC entities than other types, whereas for both Bible and Wikipedia, the models tend to perform best on PER entities.

\textit{Overall, these results highlight that model performance varies significantly across domains and entity types, with a consistent weakness in recognizing ORG entities.}

\section{Conclusion}

In this paper, we introduce \yoner, a multi-domain human annotated NER dataset \yoruba that extends entity coverage beyond news and Wikipedia. The dataset comprises about 5,000 sentences and 100,000 tokens collected from five domains including Bible, Blogs, Movies, Radio broadcast, and Wikipedia, and annotating them for person, organization, and location entities.  We benchmark several transformer encoder models using cross-domain experiments with MasakhaNER 2.0, and we also assess the effect of few-shot in-domain data using \yoner and cross-lingual setups with English datasets. Our results show that African-centric models outperform general multilingual models for \yoruba, but cross-domain performance drops substantially, particularly for blogs and movie domains. Furthermore, we observed that closely related formal domains, such as news and Wikipedia, transfer more effectively. 

\begin{table}[t]
\centering
\scalebox{0.90}{
\begin{tabular}{lccc}
\toprule
\textbf{Domain} & \textbf{PER} & \textbf{LOC} & \textbf{ORG}  \\
\midrule
\multicolumn{4}{l}{\emph{OyoBERT-large}} \\
Bible & 0.89 & 0.80 & 0.00\\
Blog &  0.70 & 0.80 & 0.42\\
Movies & 0.61 & 0.41 & 0.00\\
Radio & 0.88 & 0.95 & 0.80\\
Wiki & 0.87 & 0.84 & 0.53\\
\midrule
\multicolumn{4}{l}{\emph{AfroXLMR-large}} \\
Bible & 0.91 & 0.83 & 0.00\\
Blog &  0.66 & 0.82 & 0.57\\
Movies & 0.50 & 0.37 & 0.00\\
Radio & 0.89 & 0.96 & 0.89\\
Wiki & 0.83 & 0.84 & 0.66\\
\bottomrule
\end{tabular}
}
\caption{F1 score (\%) for multi-domain training on YoNER. Average is over 5 runs.}
\label{tab:engcrossent}
\end{table}

Given the small size of the training split in \yoner, one aspect we did not explore but which remains important is few-shot NER. However, many recent approaches to few-shot learning require a language-specific model~\citep{ding-etal-2021-nerd,das-etal-2022-container}. In this paper, we created a Yoruba-specific model in OyoBERT, and we hope that future work will build upon it and \yoner to advance few-shot NER research for \yoruba and other low-resource African languages.


\section{Limitation}
In this paper, we introduce \yoner, a moderately large multi-domain NER dataset covering five domains. Three of these domains contain little to no ORG entities due to their nature, and none include DATE entities. We hope that future work can extend \yoner to additional domains and a broader range of entity types. We did not evaluate large language models, which are now ubiquitous and general-purpose, and we hope that future work will explore their performance on \yoner.

Lastly, a limitation of this work is that the pre-trained \yoruba encoder models were trained on relatively small corpora, including translated texts. While they achieve strong performance on the evaluated downstream tasks, we do not perform an extensive analysis of their pretraining quality or representations, which would be necessary to fully understand their capabilities. Future work should therefore examine the properties captured by these representations, evaluate potential biases introduced by translated data, and explore scaling pretraining with larger and more diverse \yoruba corpora. Such investigations would provide a clearer picture of the models' generalization ability and their suitability for broader downstream applications.

\section{Acknowledgment}
David Adelani acknowledges the funding of Natural Sciences and Engineering Research Council (NSERC) of Canada,  IVADO and the Canada First Research Excellence Fund. 

\section{Bibliographical References}\label{sec:reference}
\bibliographystyle{lrec2026-natbib}
\bibliography{lrec2026-example}

\begin{thebibliography}{10}
\expandafter\ifx\csname natexlab\endcsname\relax\def\natexlab#1{#1}\fi

\bibitem[{Adelani et~al.(2021)Adelani, Abbott, Neubig, D{'}souza, Kreutzer, Lignos, Palen-Michel, Buzaaba, Rijhwani, Ruder, Mayhew, Azime, Muhammad, Emezue, Nakatumba-Nabende, Ogayo, Anuoluwapo, Gitau, Mbaye, Alabi, Yimam, Gwadabe, Ezeani, Niyongabo, Mukiibi, Otiende, Orife, David, Ngom, Adewumi, Rayson, Adeyemi, Muriuki, Anebi, Chukwuneke, Odu, Wairagala, Oyerinde, Siro, Bateesa, Oloyede, Wambui, Akinode, Nabagereka, Katusiime, Awokoya, MBOUP, Gebreyohannes, Tilaye, Nwaike, Wolde, Faye, Sibanda, Ahia, Dossou, Ogueji, DIOP, Diallo, Akinfaderin, Marengereke, and Osei}]{adelani-etal-2021-masakhaner}
Adelani, David Ifeoluwa and Abbott, Jade and Neubig, Graham and D{'}souza, Daniel and Kreutzer, Julia and Lignos, Constantine and Palen-Michel, Chester and Buzaaba, Happy and Rijhwani, Shruti and Ruder, Sebastian and Mayhew, Stephen and Azime, Israel Abebe and Muhammad, Shamsuddeen H. and Emezue, Chris Chinenye and Nakatumba-Nabende, Joyce and Ogayo, Perez and Anuoluwapo, Aremu and Gitau, Catherine and Mbaye, Derguene and Alabi, Jesujoba and Yimam, Seid Muhie and Gwadabe, Tajuddeen Rabiu and Ezeani, Ignatius and Niyongabo, Rubungo Andre and Mukiibi, Jonathan and Otiende, Verrah and Orife, Iroro and David, Davis and Ngom, Samba and Adewumi, Tosin and Rayson, Paul and Adeyemi, Mofetoluwa and Muriuki, Gerald and Anebi, Emmanuel and Chukwuneke, Chiamaka and Odu, Nkiruka and Wairagala, Eric Peter and Oyerinde, Samuel and Siro, Clemencia and Bateesa, Tobius Saul and Oloyede, Temilola and Wambui, Yvonne and Akinode, Victor and Nabagereka, Deborah and Katusiime, Maurice and Awokoya, Ayodele and MBOUP, Mouhamadane and
  Gebreyohannes, Dibora and Tilaye, Henok and Nwaike, Kelechi and Wolde, Degaga and Faye, Abdoulaye and Sibanda, Blessing and Ahia, Orevaoghene and Dossou, Bonaventure F. P. and Ogueji, Kelechi and DIOP, Thierno Ibrahima and Diallo, Abdoulaye and Akinfaderin, Adewale and Marengereke, Tendai and Osei, Salomey. 2021.
\newblock \href {https://doi.org/10.1162/tacl_a_00416} {\emph{{M}asakha{NER}: Named Entity Recognition for {A}frican Languages}}.
\newblock MIT Press.

\bibitem[{Adelani et~al.(2024)Adelani, Liu, Shen, Vassilyev, Alabi, Mao, Gao, and Lee}]{adelani-etal-2024-sib}
Adelani, David Ifeoluwa and Liu, Hannah and Shen, Xiaoyu and Vassilyev, Nikita and Alabi, Jesujoba O. and Mao, Yanke and Gao, Haonan and Lee, En-Shiun Annie. 2024.
\newblock \href {https://doi.org/10.18653/v1/2024.eacl-long.14} {\emph{{SIB}-200: A Simple, Inclusive, and Big Evaluation Dataset for Topic Classification in 200+ Languages and Dialects}}.
\newblock Association for Computational Linguistics.

\bibitem[{Adelani et~al.(2022)Adelani, Neubig, Ruder, Rijhwani, Beukman, Palen-Michel, Lignos, Alabi, Muhammad, Nabende, Dione, Bukula, Mabuya, Dossou, Sibanda, Buzaaba, Mukiibi, Kalipe, Mbaye, Taylor, Kabore, Emezue, Aremu, Ogayo, Gitau, Munkoh-Buabeng, Memdjokam~Koagne, Tapo, Macucwa, Marivate, Mboning, Gwadabe, Adewumi, Ahia, Nakatumba-Nabende, Mokono, Ezeani, Chukwuneke, Adeyemi, Hacheme, Abdulmumin, Ogundepo, Yousuf, Moteu~Ngoli, and Klakow}]{adelani-etal-2022-masakhaner}
Adelani, David Ifeoluwa and Neubig, Graham and Ruder, Sebastian and Rijhwani, Shruti and Beukman, Michael and Palen-Michel, Chester and Lignos, Constantine and Alabi, Jesujoba O. and Muhammad, Shamsuddeen H. and Nabende, Peter and Dione, Cheikh M. Bamba and Bukula, Andiswa and Mabuya, Rooweither and Dossou, Bonaventure F. P. and Sibanda, Blessing and Buzaaba, Happy and Mukiibi, Jonathan and Kalipe, Godson and Mbaye, Derguene and Taylor, Amelia and Kabore, Fatoumata and Emezue, Chris Chinenye and Aremu, Anuoluwapo and Ogayo, Perez and Gitau, Catherine and Munkoh-Buabeng, Edwin and Memdjokam Koagne, Victoire and Tapo, Allahsera Auguste and Macucwa, Tebogo and Marivate, Vukosi and Mboning, Elvis and Gwadabe, Tajuddeen and Adewumi, Tosin and Ahia, Orevaoghene and Nakatumba-Nabende, Joyce and Mokono, Neo L. and Ezeani, Ignatius and Chukwuneke, Chiamaka and Adeyemi, Mofetoluwa and Hacheme, Gilles Q. and Abdulmumin, Idris and Ogundepo, Odunayo and Yousuf, Oreen and Moteu Ngoli, Tatiana and Klakow, Dietrich. 2022.
\newblock \href {https://doi.org/10.18653/v1/2022.emnlp-main.298} {\emph{{M}asakha{NER} 2.0: {A}frica-centric Transfer Learning for Named Entity Recognition}}.
\newblock Association for Computational Linguistics.

\bibitem[{Alabi et~al.(2022)Alabi, Adelani, Mosbach, and Klakow}]{alabi2022adapting}
Alabi, Jesujoba and Adelani, David Ifeoluwa and Mosbach, Marius and Klakow, Dietrich. 2022.
\newblock \emph{Adapting Pre-trained Language Models to African Languages via Multilingual Adaptive Fine-Tuning}.

\bibitem[{Conneau et~al.(2020)Conneau, Khandelwal, Goyal, Chaudhary, Wenzek, Guzm{\'a}n, Grave, Ott, Zettlemoyer, and Stoyanov}]{conneau-etal-2020-unsupervised}
Conneau, Alexis and Khandelwal, Kartikay and Goyal, Naman and Chaudhary, Vishrav and Wenzek, Guillaume and Guzm{\'a}n, Francisco and Grave, Edouard and Ott, Myle and Zettlemoyer, Luke and Stoyanov, Veselin. 2020.
\newblock \href {https://doi.org/10.18653/v1/2020.acl-main.747} {\emph{Unsupervised Cross-lingual Representation Learning at Scale}}.
\newblock Association for Computational Linguistics.

\bibitem[{Devlin et~al.(2019)Devlin, Chang, Lee, and Toutanova}]{devlin-etal-2019-bert}
Devlin, Jacob and Chang, Ming-Wei and Lee, Kenton and Toutanova, Kristina. 2019.
\newblock \href {https://doi.org/10.18653/v1/N19-1423} {\emph{{BERT}: Pre-training of Deep Bidirectional Transformers for Language Understanding}}.
\newblock Association for Computational Linguistics.

\bibitem[{Ogueji et~al.(2021)Ogueji, Zhu, and Lin}]{ogueji-etal-2021-small}
Ogueji, Kelechi and Zhu, Yuxin and Lin, Jimmy. 2021.
\newblock \href {https://doi.org/10.18653/v1/2021.mrl-1.11} {\emph{Small Data? No Problem! Exploring the Viability of Pretrained Multilingual Language Models for Low-resourced Languages}}.
\newblock Association for Computational Linguistics.

\bibitem[{Oladipo(2024)}]{oladipo2024scaling}
Oladipo, Akintunde. 2024.
\newblock \emph{Scaling pre-training data and language models for african languages}.
\newblock University of Waterloo.

\bibitem[{Oyewusi et~al.(2021)Oyewusi, Adekanmbi, Okoh, Onuigwe, Salami, Osakuade, Ibejih, and Musa}]{oyewusi2021naijaner}
Wuraola~Fisayo Oyewusi, Olubayo Adekanmbi, Ifeoma Okoh, Vitus Onuigwe, Mary~Idera Salami, Opeyemi Osakuade, Sharon Ibejih, and Usman~Abdullahi Musa. 2021.
\newblock Naijaner: Comprehensive named entity recognition for 5 nigerian languages.
\newblock \emph{arXiv preprint arXiv:2105.00810}.

\bibitem[{Pan et~al.(2017)Pan, Zhang, May, Nothman, Knight, and Ji}]{pan-etal-2017-cross}
Pan, Xiaoman and Zhang, Boliang and May, Jonathan and Nothman, Joel and Knight, Kevin and Ji, Heng. 2017.
\newblock \href {https://doi.org/10.18653/v1/P17-1178} {\emph{Cross-lingual Name Tagging and Linking for 282 Languages}}.
\newblock Association for Computational Linguistics.

\end{thebibliography}


\begin{thebibliography}{35}
\expandafter\ifx\csname natexlab\endcsname\relax\def\natexlab#1{#1}\fi

\bibitem[{Adelani et~al.(2021)Adelani, Abbott, Neubig, D’souza, Kreutzer, Lignos, Osei et~al.}]{adelani2021masakhane}
David~Ifeoluwa Adelani, Jade Abbott, Graham Neubig, Daniel D’souza, Julia Kreutzer, Constantine Lignos, Selasi Osei, et~al. 2021.
\newblock Masakhaner: Named entity recognition for african languages.
\newblock \emph{Transactions of the Association for Computational Linguistics}, 9:1116--1131.

\bibitem[{Adelani et~al.(2024)Adelani, Liu, Shen, Vassilyev, Alabi, Mao, Gao, and Lee}]{adelani2024sib}
David~Ifeoluwa Adelani, Hannah Liu, Xiaoyu Shen, Nikita Vassilyev, Jesujoba Alabi, Yanke Mao, Haonan Gao, and En-Shiun~Annie Lee. 2024.
\newblock Sib-200: A simple, inclusive, and big evaluation dataset for topic classification in 200+ languages and dialects.
\newblock In \emph{Proceedings of the 18th Conference of the European Chapter of the Association for Computational Linguistics (Volume 1: Long Papers)}, pages 226--245.

\bibitem[{Adelani et~al.(2023)Adelani, Masiak, Azime, Alabi, Tonja, Mwase, Ogundepo, Dossou, Oladipo, Nixdorf et~al.}]{adelani2023masakhanews}
David~Ifeoluwa Adelani, Marek Masiak, Israel~Abebe Azime, Jesujoba Alabi, Atnafu~Lambebo Tonja, Christine Mwase, Odunayo Ogundepo, Bonaventure~FP Dossou, Akintunde Oladipo, Doreen Nixdorf, et~al. 2023.
\newblock Masakhanews: News topic classification for african languages.
\newblock In \emph{Proceedings of the 13th International Joint Conference on Natural Language Processing and the 3rd Conference of the Asia-Pacific Chapter of the Association for Computational Linguistics (Volume 1: Long Papers)}, pages 144--159.

\bibitem[{Adelani et~al.(2022)Adelani, Neubig, Ruder, Rijhwani, Beukman, Palen-Michel, Lignos, Alabi, Muhammad, Nabende, Dione, Bukula, Mabuya, Dossou, Sibanda, Buzaaba, Mukiibi, Kalipe, Mbaye, Taylor, Kabore, Emezue, Aremu, Ogayo, Gitau, Munkoh-Buabeng, Memdjokam~Koagne, Tapo, Macucwa, Marivate, Mboning, Gwadabe, Adewumi, Ahia, Nakatumba-Nabende, Mokono, Ezeani, Chukwuneke, Adeyemi, Hacheme, Abdulmumin, Ogundepo, Yousuf, Moteu~Ngoli, and Klakow}]{adelani-etal-2022-masakhaner}
David~Ifeoluwa Adelani, Graham Neubig, Sebastian Ruder, Shruti Rijhwani, Michael Beukman, Chester Palen-Michel, Constantine Lignos, Jesujoba~O. Alabi, Shamsuddeen~H. Muhammad, Peter Nabende, Cheikh M.~Bamba Dione, Andiswa Bukula, Rooweither Mabuya, Bonaventure F.~P. Dossou, Blessing Sibanda, Happy Buzaaba, Jonathan Mukiibi, Godson Kalipe, Derguene Mbaye, Amelia Taylor, Fatoumata Kabore, Chris~Chinenye Emezue, Anuoluwapo Aremu, Perez Ogayo, Catherine Gitau, Edwin Munkoh-Buabeng, Victoire Memdjokam~Koagne, Allahsera~Auguste Tapo, Tebogo Macucwa, Vukosi Marivate, Elvis Mboning, Tajuddeen Gwadabe, Tosin Adewumi, Orevaoghene Ahia, Joyce Nakatumba-Nabende, Neo~L. Mokono, Ignatius Ezeani, Chiamaka Chukwuneke, Mofetoluwa Adeyemi, Gilles~Q. Hacheme, Idris Abdulmumin, Odunayo Ogundepo, Oreen Yousuf, Tatiana Moteu~Ngoli, and Dietrich Klakow. 2022.
\newblock \href {https://doi.org/10.18653/v1/2022.emnlp-main.298} {{M}asakha{NER} 2.0: {A}frica-centric transfer learning for named entity recognition}.
\newblock In \emph{Proceedings of the 2022 Conference on Empirical Methods in Natural Language Processing}, pages 4488--4508, Abu Dhabi, United Arab Emirates. Association for Computational Linguistics.

\bibitem[{Alabi et~al.(2022)Alabi, Adelani, Mosbach, and Klakow}]{alabi2022adapting}
Jesujoba Alabi, David~Ifeoluwa Adelani, Marius Mosbach, and Dietrich Klakow. 2022.
\newblock Adapting pre-trained language models to african languages via multilingual adaptive fine-tuning.
\newblock In \emph{Proceedings of the 29th International Conference on Computational Linguistics}, pages 4336--4349.

\bibitem[{Alabi et~al.(2020)Alabi, Amponsah-Kaakyire, Adelani, and Espa{\~n}a-Bonet}]{alabi2020massive}
Jesujoba Alabi, Kwabena Amponsah-Kaakyire, David~Ifeoluwa Adelani, and Cristina Espa{\~n}a-Bonet. 2020.
\newblock Massive vs. curated embeddings for low-resourced languages: The case of yor{\`u}b{\'a} and twi.
\newblock In \emph{Proceedings of the Twelfth Language Resources and Evaluation Conference (LREC)}, pages 2754--2762, Marseille, France.

\bibitem[{Ashok and Lipton(2023)}]{ashok2023promptner}
Dhananjay Ashok and Zachary~C Lipton. 2023.
\newblock Promptner: Prompting for named entity recognition.
\newblock \emph{arXiv preprint arXiv:2305.15444}.

\bibitem[{Caubri{\`e}re et~al.(2020)Caubri{\`e}re, Rosset, Est{\`e}ve, Laurent, and Morin}]{caubriere-etal-2020-named}
Antoine Caubri{\`e}re, Sophie Rosset, Yannick Est{\`e}ve, Antoine Laurent, and Emmanuel Morin. 2020.
\newblock \href {https://aclanthology.org/2020.lrec-1.556/} {Where are we in named entity recognition from speech?}
\newblock In \emph{Proceedings of the Twelfth Language Resources and Evaluation Conference}, pages 4514--4520, Marseille, France. European Language Resources Association.

\bibitem[{Das et~al.(2022{\natexlab{a}})Das, Katiyar, Passonneau, and Zhang}]{das-etal-2022-container}
Sarkar Snigdha~Sarathi Das, Arzoo Katiyar, Rebecca Passonneau, and Rui Zhang. 2022{\natexlab{a}}.
\newblock \href {https://doi.org/10.18653/v1/2022.acl-long.439} {{CONT}ai{NER}: Few-shot named entity recognition via contrastive learning}.
\newblock In \emph{Proceedings of the 60th Annual Meeting of the Association for Computational Linguistics (Volume 1: Long Papers)}, pages 6338--6353, Dublin, Ireland. Association for Computational Linguistics.

\bibitem[{Das et~al.(2022{\natexlab{b}})Das, Katiyar, Passonneau, and Zhang}]{das2022container}
Sarkar Snigdha~Sarathi Das, Arzoo Katiyar, Rebecca~J Passonneau, and Rui Zhang. 2022{\natexlab{b}}.
\newblock Container: Few-shot named entity recognition via contrastive learning.
\newblock In \emph{Proceedings of the 60th Annual Meeting of the Association for Computational Linguistics (Volume 1: Long Papers)}, pages 6338--6353.

\bibitem[{Ding et~al.(2021)Ding, Xu, Chen, Wang, Han, Xie, Zheng, and Liu}]{ding-etal-2021-nerd}
Ning Ding, Guangwei Xu, Yulin Chen, Xiaobin Wang, Xu~Han, Pengjun Xie, Haitao Zheng, and Zhiyuan Liu. 2021.
\newblock \href {https://doi.org/10.18653/v1/2021.acl-long.248} {Few-{NERD}: A few-shot named entity recognition dataset}.
\newblock In \emph{Proceedings of the 59th Annual Meeting of the Association for Computational Linguistics and the 11th International Joint Conference on Natural Language Processing (Volume 1: Long Papers)}, pages 3198--3213, Online. Association for Computational Linguistics.

\bibitem[{Dione et~al.(2023)Dione, Adelani, Nabende, Alabi, Sindane, Buzaaba, Muhammad, Emezue, Ogayo, Aremu et~al.}]{dione2023masakhapos}
Cheikh M~Bamba Dione, David~Ifeoluwa Adelani, Peter Nabende, Jesujoba Alabi, Thapelo Sindane, Happy Buzaaba, Shamsuddeen~Hassan Muhammad, Chris~Chinenye Emezue, Perez Ogayo, Anuoluwapo Aremu, et~al. 2023.
\newblock Masakhapos: Part-of-speech tagging for typologically diverse african languages.
\newblock In \emph{Proceedings of the 61st Annual Meeting of the Association for Computational Linguistics (Volume 1: Long Papers)}, pages 10883--10900.

\bibitem[{Gowda et~al.(2021)Gowda, Zhang, Mattmann, and May}]{gowda-etal-2021-many}
Thamme Gowda, Zhao Zhang, Chris Mattmann, and Jonathan May. 2021.
\newblock \href {https://doi.org/10.18653/v1/2021.acl-demo.37} {Many-to-{E}nglish machine translation tools, data, and pretrained models}.
\newblock In \emph{Proceedings of the 59th Annual Meeting of the Association for Computational Linguistics and the 11th International Joint Conference on Natural Language Processing: System Demonstrations}, pages 306--316, Online. Association for Computational Linguistics.

\bibitem[{Hou et~al.(2022)Hou, Pang, Zhou, Wu, Song, Song, and Zhou}]{hou-etal-2022-token}
Le~Hou, Richard~Yuanzhe Pang, Tianyi Zhou, Yuexin Wu, Xinying Song, Xiaodan Song, and Denny Zhou. 2022.
\newblock \href {https://doi.org/10.18653/v1/2022.acl-long.262} {Token dropping for efficient {BERT} pretraining}.
\newblock In \emph{Proceedings of the 60th Annual Meeting of the Association for Computational Linguistics (Volume 1: Long Papers)}, pages 3774--3784, Dublin, Ireland. Association for Computational Linguistics.

\bibitem[{Huang et~al.(2025)Huang, Liu, Gao, Yu, Liu, and Chen}]{huang2025adversity}
Li~Huang, Haowen Liu, Qiang Gao, Jiajing Yu, Guisong Liu, and Xueqin Chen. 2025.
\newblock Adversity-aware few-shot named entity recognition via augmentation learning.
\newblock In \emph{Proceedings of the AAAI Conference on Artificial Intelligence}, volume~39, pages 24132--24140.

\bibitem[{Jia et~al.(2019)Jia, Liang, and Zhang}]{jia-etal-2019-cross}
Chen Jia, Xiaobo Liang, and Yue Zhang. 2019.
\newblock \href {https://doi.org/10.18653/v1/P19-1236} {Cross-domain {NER} using cross-domain language modeling}.
\newblock In \emph{Proceedings of the 57th Annual Meeting of the Association for Computational Linguistics}, pages 2464--2474, Florence, Italy. Association for Computational Linguistics.

\bibitem[{Liu et~al.(2021)Liu, Xu, Yu, Dai, Ji, Cahyawijaya, Fung et~al.}]{liu2021crossner}
Zhengbao Liu, Yichong Xu, Tianyi Yu, Wenlong Dai, Ziwei Ji, Samuel Cahyawijaya, Pascale Fung, et~al. 2021.
\newblock Crossner: Evaluating cross-domain named entity recognition.
\newblock In \emph{Proceedings of the AAAI Conference on Artificial Intelligence}, volume~35, pages 13452--13460, Virtual.

\bibitem[{Muhammad et~al.(2023)Muhammad, Abdulmumin, Ayele, Ousidhoum, Adelani, Yimam, Ahmad, Beloucif, Mohammad, Ruder et~al.}]{muhammad2023afrisenti}
Shamsuddeen~Hassan Muhammad, Idris Abdulmumin, Abinew~Ali Ayele, Nedjma Ousidhoum, David~Ifeoluwa Adelani, Seid~Muhie Yimam, Ibrahim~Sa'id Ahmad, Meriem Beloucif, Saif Mohammad, Sebastian Ruder, et~al. 2023.
\newblock Afrisenti: A twitter sentiment analysis benchmark for african languages.
\newblock In \emph{Proceedings of the 2023 Conference on Empirical Methods in Natural Language Processing}, pages 13968--13981.

\bibitem[{NLLB-Team et~al.(2022)NLLB-Team, Costa-Juss{\`a}, Cross, {\c{C}}elebi, Elbayad, Heafield, Heffernan, Kalbassi, Lam, Licht, Maillard et~al.}]{costa2022no}
NLLB-Team, Marta~R Costa-Juss{\`a}, James Cross, Onur {\c{C}}elebi, Maha Elbayad, Kenneth Heafield, Kevin Heffernan, Elahe Kalbassi, Janice Lam, Daniel Licht, Jean Maillard, et~al. 2022.
\newblock No language left behind: Scaling human-centered machine translation.
\newblock \emph{arXiv preprint arXiv:2207.04672}.

\bibitem[{{NLLB Team} et~al.(2024)}]{nllb2024scaling}
{NLLB Team} et~al. 2024.
\newblock Scaling neural machine translation to 200 languages.
\newblock \emph{Nature}, 630(8018):841.

\bibitem[{Ogueji et~al.(2021)Ogueji, Zhu, and Lin}]{ogueji-etal-2021-small}
Kelechi Ogueji, Yuxin Zhu, and Jimmy Lin. 2021.
\newblock \href {https://doi.org/10.18653/v1/2021.mrl-1.11} {Small data? no problem! exploring the viability of pretrained multilingual language models for low-resourced languages}.
\newblock In \emph{Proceedings of the 1st Workshop on Multilingual Representation Learning}, pages 116--126, Punta Cana, Dominican Republic. Association for Computational Linguistics.

\bibitem[{Oladipo et~al.(2023)Oladipo, Adeyemi, Ahia, Owodunni, Ogundepo, Adelani, and Lin}]{oladipo-etal-2023-better}
Akintunde Oladipo, Mofetoluwa Adeyemi, Orevaoghene Ahia, Abraham~Toluwalase Owodunni, Odunayo Ogundepo, David~Ifeoluwa Adelani, and Jimmy Lin. 2023.
\newblock \href {https://doi.org/10.18653/v1/2023.emnlp-main.11} {Better quality pre-training data and t5 models for {A}frican languages}.
\newblock In \emph{Proceedings of the 2023 Conference on Empirical Methods in Natural Language Processing}, pages 158--168, Singapore. Association for Computational Linguistics.

\bibitem[{Olatunji et~al.(2023)Olatunji, Afonja, Dossou, Tonja, Emezue, Rufai, and Singh}]{olatunji2023afrinames}
Tobi Olatunji, Tejumade Afonja, Bonaventure~FP Dossou, Atnafu~Lambebo Tonja, Chris~Chinenye Emezue, Amina~Mardiyyah Rufai, and Sahib Singh. 2023.
\newblock Afrinames: Most asr models.
\newblock In \emph{Proc. Interspeech 2023}, pages 5077--5081.

\bibitem[{Palen-Michel et~al.(2024)Palen-Michel, Pickering, Kruse, S{\"a}lev{\"a}, and Lignos}]{palen2024openner}
Chester Palen-Michel, Maxwell Pickering, Maya Kruse, Jonne S{\"a}lev{\"a}, and Constantine Lignos. 2024.
\newblock Openner 1.0: Standardized open-access named entity recognition datasets in 50+ languages.
\newblock \emph{arXiv preprint arXiv:2412.09587}.

\bibitem[{Pyysalo et~al.(2021)Pyysalo, Kanerva, Virtanen, and Ginter}]{pyysalo-etal-2021-wikibert}
Sampo Pyysalo, Jenna Kanerva, Antti Virtanen, and Filip Ginter. 2021.
\newblock \href {https://aclanthology.org/2021.nodalida-main.1/} {{W}iki{BERT} models: Deep transfer learning for many languages}.
\newblock In \emph{Proceedings of the 23rd Nordic Conference on Computational Linguistics (NoDaLiDa)}, pages 1--10, Reykjavik, Iceland (Online). Link{\"o}ping University Electronic Press, Sweden.

\bibitem[{R{\"o}nnqvist et~al.(2019)R{\"o}nnqvist, Kanerva, Salakoski, and Ginter}]{ronnqvist-etal-2019-multilingual}
Samuel R{\"o}nnqvist, Jenna Kanerva, Tapio Salakoski, and Filip Ginter. 2019.
\newblock \href {https://aclanthology.org/W19-6204/} {Is multilingual {BERT} fluent in language generation?}
\newblock In \emph{Proceedings of the First NLPL Workshop on Deep Learning for Natural Language Processing}, pages 29--36, Turku, Finland. Link{\"o}ping University Electronic Press.

\bibitem[{Schmidt et~al.(2022)Schmidt, Vuli{\'c}, and Glava{\v{s}}}]{schmidt-etal-2022-dont}
Fabian~David Schmidt, Ivan Vuli{\'c}, and Goran Glava{\v{s}}. 2022.
\newblock \href {https://doi.org/10.18653/v1/2022.emnlp-main.736} {Don{'}t stop fine-tuning: On training regimes for few-shot cross-lingual transfer with multilingual language models}.
\newblock In \emph{Proceedings of the 2022 Conference on Empirical Methods in Natural Language Processing}, pages 10725--10742, Abu Dhabi, United Arab Emirates. Association for Computational Linguistics.

\bibitem[{Shode et~al.(2023)Shode, Adelani, Peng, and Feldman}]{shode2023nollysenti}
Iyanuoluwa Shode, David~Ifeoluwa Adelani, Jing Peng, and Anna Feldman. 2023.
\newblock Nollysenti: Leveraging transfer learning and machine translation for nigerian movie sentiment classification.
\newblock In \emph{Proceedings of the 61st Annual Meeting of the Association for Computational Linguistics (Volume 2: Short Papers)}, pages 986--998.

\bibitem[{Tjong Kim~Sang and De~Meulder(2003)}]{tjong-kim-sang-de-meulder-2003-introduction}
Erik~F. Tjong Kim~Sang and Fien De~Meulder. 2003.
\newblock \href {https://aclanthology.org/W03-0419/} {Introduction to the {C}o{NLL}-2003 shared task: Language-independent named entity recognition}.
\newblock In \emph{Proceedings of the Seventh Conference on Natural Language Learning at {HLT}-{NAACL} 2003}, pages 142--147.

\bibitem[{Weischedel et~al.(2011)Weischedel, Hovy, Marcus, Palmer, Belvin, Pradhan, Ramshaw, and Xue}]{weischedel2011ontonotes}
Ralph Weischedel, Eduard Hovy, Mitchell Marcus, Martha Palmer, Robert Belvin, Sameer Pradhan, Lance Ramshaw, and Nianwen Xue. 2011.
\newblock Ontonotes: A large training corpus for enhanced processing.
\newblock \emph{Handbook of Natural Language Processing and Machine Translation. Springer}, 3(3):3--4.

\bibitem[{Xue et~al.(2021)Xue, Constant, Roberts, Kale, Al-Rfou, Siddhant, Barua, and Raffel}]{xue-etal-2021-mt5}
Linting Xue, Noah Constant, Adam Roberts, Mihir Kale, Rami Al-Rfou, Aditya Siddhant, Aditya Barua, and Colin Raffel. 2021.
\newblock \href {https://doi.org/10.18653/v1/2021.naacl-main.41} {m{T}5: A massively multilingual pre-trained text-to-text transformer}.
\newblock In \emph{Proceedings of the 2021 Conference of the North American Chapter of the Association for Computational Linguistics: Human Language Technologies}, pages 483--498, Online. Association for Computational Linguistics.

\bibitem[{Xue et~al.(2024)Xue, Zhang, Xu, and Niu}]{xue2024robust}
Xiaojun Xue, Chunxia Zhang, Tianxiang Xu, and Zhendong Niu. 2024.
\newblock Robust few-shot named entity recognition with boundary discrimination and correlation purification.
\newblock In \emph{Proceedings of the AAAI Conference on Artificial Intelligence}, volume~38, pages 19341--19349.

\bibitem[{Yamada et~al.(2020)Yamada, Asai, Shindo, Takeda, and Matsumoto}]{yamada-etal-2020-luke}
Ikuya Yamada, Akari Asai, Hiroyuki Shindo, Hideaki Takeda, and Yuji Matsumoto. 2020.
\newblock \href {https://doi.org/10.18653/v1/2020.emnlp-main.523} {{LUKE}: Deep contextualized entity representations with entity-aware self-attention}.
\newblock In \emph{Proceedings of the 2020 Conference on Empirical Methods in Natural Language Processing (EMNLP)}, pages 6442--6454, Online. Association for Computational Linguistics.

\bibitem[{Yang and Katiyar(2020)}]{yang2020simple}
Yi~Yang and Arzoo Katiyar. 2020.
\newblock Simple and effective few-shot named entity recognition with structured nearest neighbor learning.
\newblock In \emph{Proceedings of the 2020 Conference on Empirical Methods in Natural Language Processing (EMNLP)}, pages 6365--6375.

\bibitem[{Zhao et~al.(2021)Zhao, Zhu, Shareghi, Vuli{\'c}, Reichart, Korhonen, and Sch{\"u}tze}]{zhao-etal-2021-closer}
Mengjie Zhao, Yi~Zhu, Ehsan Shareghi, Ivan Vuli{\'c}, Roi Reichart, Anna Korhonen, and Hinrich Sch{\"u}tze. 2021.
\newblock \href {https://doi.org/10.18653/v1/2021.acl-long.447} {A closer look at few-shot crosslingual transfer: The choice of shots matters}.
\newblock In \emph{Proceedings of the 59th Annual Meeting of the Association for Computational Linguistics and the 11th International Joint Conference on Natural Language Processing (Volume 1: Long Papers)}, pages 5751--5767, Online. Association for Computational Linguistics.

\end{thebibliography}

\section{Language Resource References}
\label{lr:ref}
\bibliographystylelanguageresource{lrec2026-natbib}
\bibliographylanguageresource{languageresource}

\end{document}